%% file: main.tex
\documentclass[11pt]{article}
\usepackage{graphicx}
\usepackage{eso-pic}
\input{preamble}

\AddToShipoutPictureFG*{%
  \AtPageUpperLeft{%
    \put(\LenToUnit{2.5cm},\LenToUnit{-2.5cm}){%
      \includegraphics[width=6cm]{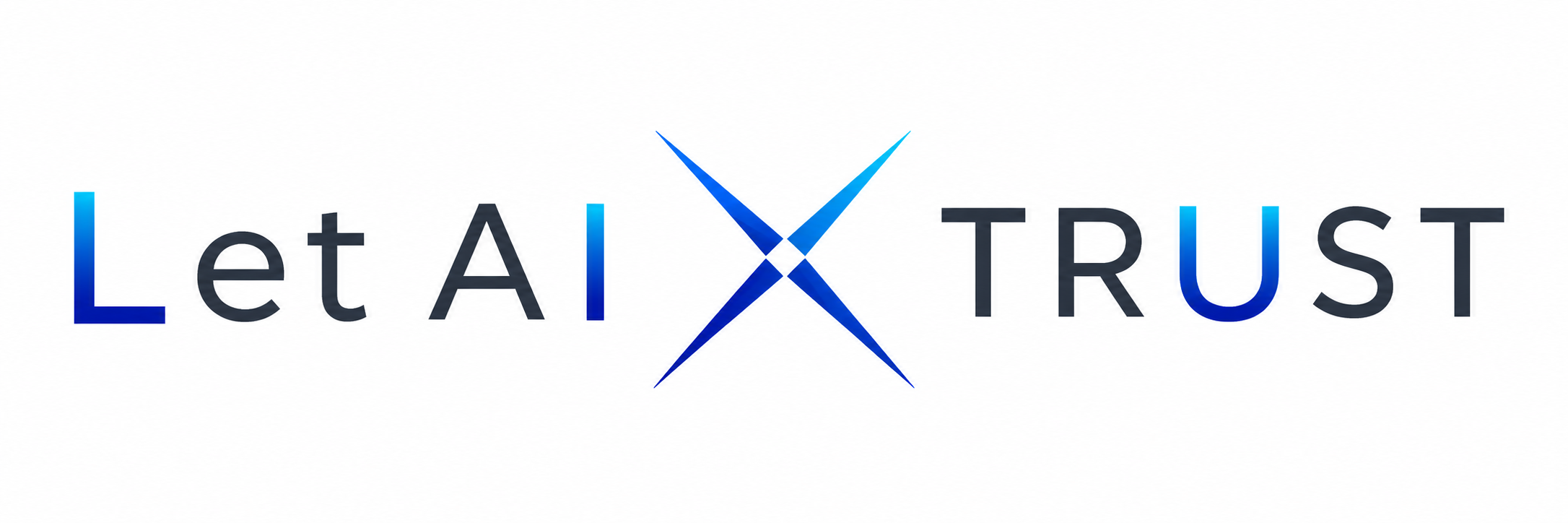}%
    }%
  }%
}

\title{Understanding Trajectory Heterogeneity in\\Federated World Model Learning}
\author{Yipan Wei$^{1}$, Zhaokun Yan$^{2}$, Ziming Hong$^{3}$, Jiaqi Wu$^{4}$, Lixu Wang$^{5,*}$
\\[5pt]
$^{1}$ Wuhan University \\
$^{2}$ China Academy of Information and Communications Technology \\
$^{3}$ The University of Sydney \\
$^{4}$ Tsinghua University \\
$^{5}$ The Chinese University of Hong Kong, Shenzhen \\[2pt]
$^{*}$Corresponding author}
\date{}
\begin{document}


\maketitle

\input{section/abstract}
\input{section/introduction}
\input{section/related_work}
\input{section/method}
\input{section/experiments}
\input{section/conclusion}

\bibliographystyle{unsrtnat}
\bibliography{references}
\end{document}

%% file: preamble.tex
\usepackage[margin=0.9in]{geometry}
\usepackage[T1]{fontenc}
\usepackage{lmodern,microtype,amsmath,amssymb,booktabs,array,graphicx}
\usepackage[numbers,sort&compress]{natbib}
\usepackage[table]{xcolor}
\usepackage{placeins,adjustbox,multirow}
\usepackage[font=small,labelfont=bf]{caption}
\definecolor{grouphead}{HTML}{F1F3F5}
\definecolor{alternaterow}{HTML}{F7F7F7}
\definecolor{baselinerow}{HTML}{FFF8D6}

\newcommand{\papertablesetup}{\setlength{\tabcolsep}{5.5pt}\setlength{\arrayrulewidth}{0.6pt}\renewcommand{\arraystretch}{1.15}}
\newcommand{\papertoprule}{\noalign{\hrule height 1.25pt}}
\newcommand{\paperbottomrule}{\noalign{\hrule height 1.25pt}}
\newcommand{\papergrouprule}{\hline\hline}

\newcommand{\paperheadrow}{\rowcolor{grouphead}}
\newcommand{\paperaltrow}{\rowcolor{alternaterow}}
\newcommand{\paperbaselinerow}{\rowcolor{baselinerow}}
\usepackage[colorlinks=true,linkcolor=blue!45!black,citecolor=blue!45!black,urlcolor=blue!45!black]{hyperref}
\usepackage{url}
\usepackage{pifont}
\definecolor{coveragegreen}{HTML}{247A59}
\definecolor{coveragemuted}{HTML}{7D8790}
\definecolor{coverageours}{HTML}{EAF4EF}
\newcommand{\coverageyes}{\textcolor{coveragegreen}{\ding{51}}}
\newcommand{\coverageno}{\textcolor{coveragemuted}{\textendash}}

%% file: section/abstract.tex
\begin{abstract}
World models learn state evolution from trajectories, making access to temporal context a central training requirement. Federated learning can use distributed records, while ownership boundaries within a trajectory restrict the examples each client can construct. Our study benchmarks this cross-time setting through hourly action-conditioned clinical prediction on eight MIMIC-IV disease cohorts, comprising 40.87 million transition memberships. We specify severity-based client ownership, patient-separated construction, local history and future-window rules, and paired rollout evaluation from one to 32 hours. A matrix of ten federated algorithms covers 32 disease--partition configurations under five rounds of ten-percent participation. Three findings emerge from existing results and training logs. First, client ownership and participation jointly restrict long-window coverage: only 7.55\%--21.36\% of pooled-available 32-step windows have a locally complete anchor visited during training, averaged across diseases. Second, finer severity partitions accompany higher FedAvg error in 15 of 16 paired comparisons, while algorithm gains are small and horizon-dependent: FedProx reduces mean error by 0.56\%, with no consistent improvement at 32 steps. Third, algorithm labels conceal distinct update behavior, including inactive extrapolation and orders-of-magnitude differences in update scale. Cached-update performance also varies strongly across trajectory partitions under the same benchmark protocol. These results establish temporal access, participation coverage, optimization behavior, and horizon-resolved prediction as complementary dimensions for evaluating federated clinical world models.
\end{abstract}

%% file: section/introduction.tex
\section{Introduction}
A world model learns how an environment evolves and predicts future states conditional on actions. Learned latent dynamics support planning and behavior learning in World Models, PlaNet, and Dreamer~\cite{ha2018worldmodels,hafner2019planet,hafner2025dreamer}. In a clinical trajectory, the corresponding predictive problem is to summarize a patient's recent physiological history and predict subsequent states along a recorded treatment path. This example makes the learning requirement concrete: the model must connect what happened before an action with the states observed afterward.

The data requirement therefore has three parts: enough trajectories to learn recurring dynamics, diverse conditions under which those dynamics occur, and sufficient temporal context to learn their evolution. A collection of isolated observations cannot supply the same training examples as ordered histories and future windows. This motivates learning across distributed trajectory collections, where additional clients can contribute experience beyond one local dataset.

\begin{figure}[t]
\centering
\includegraphics[width=\textwidth]{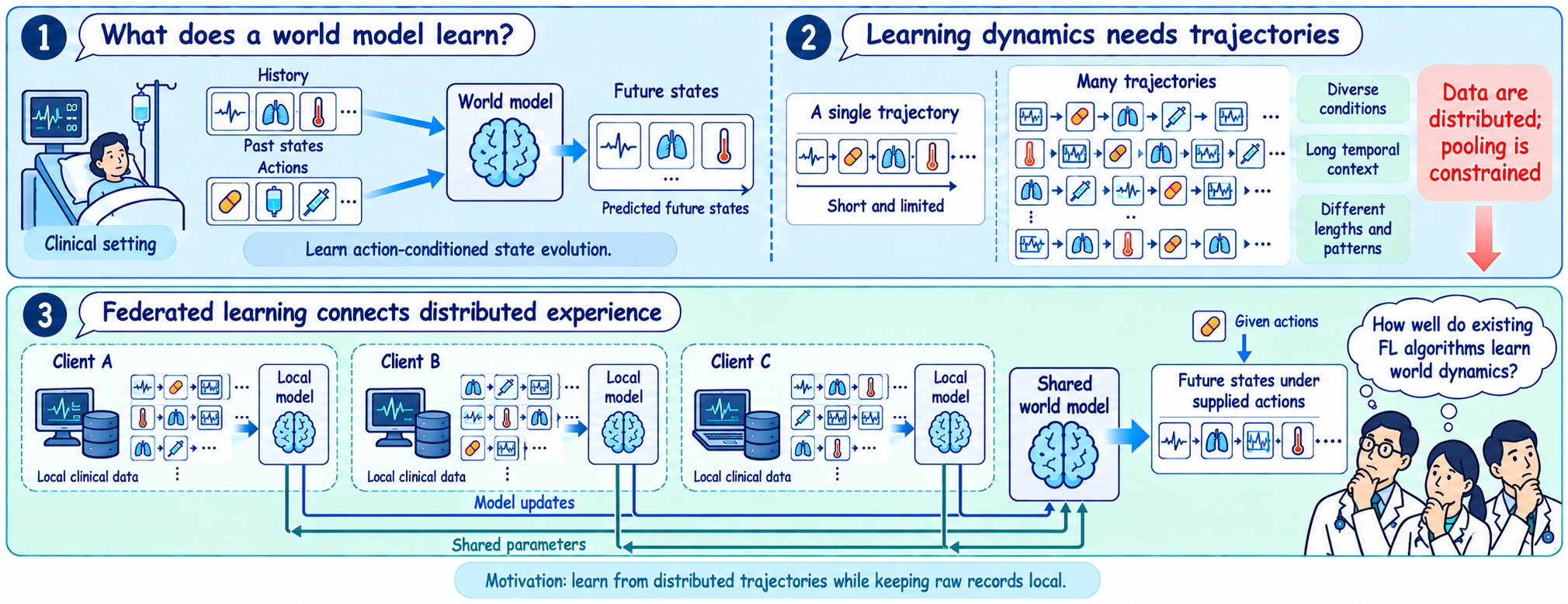}
\caption{World models motivate federated learning from distributed trajectories. History and actions condition future-state prediction; learning these dynamics draws on trajectory breadth, diversity, and temporal context. FL connects local learners through model updates while records remain local.}
\label{fig:motivation}
\end{figure}

Federated learning (FL) provides a mechanism for using this distributed experience: clients train locally and exchange model updates while retaining raw records~\cite{mcmahan2017fedavg,rieke2020digitalhealthfl}. Figure~\ref{fig:motivation} summarizes the connection between world-model learning and federated collaboration. The resulting research question is whether existing FL algorithms can turn access to more distributed records into useful trajectory learning. Answering it requires specifying which parts of a trajectory each client can actually use.

We study \emph{cross-time federated world modeling}, in which distinct time segments of the same trajectory can belong to different clients. Consider a patient's condition worsening and later improving: stage-based ownership can give one client an earlier and a later segment while assigning the intervening segment elsewhere. A local learner then has partial historical context and a limited future training window. We instantiate this setting through simulated severity-stage clients within MIMIC-IV. Each disease is a separately trained task; the experiments evaluate temporal ownership within that task.

Existing federated benchmarks provide complementary foundations. LEAF emphasizes federated data and evaluation, FedScale combines workloads with system-level evaluation, and FLamby provides healthcare tasks with natural institutional splits~\cite{caldas2018leaf,lai2022fedscale,terrail2022flamby}. Related time-series work studies cross-device heterogeneity and continual forecasting, while FedTDP explicitly partitions mobility trajectories across regional clients for data preparation~\cite{yuan2024fedtrend,hallak2025forgetting,zeng2026fedtdp}. Our study focuses on action-conditioned clinical prediction under severity-based temporal ownership. Its evaluation connects the windows available to local learners with the participation schedule, actual optimization behavior, and multi-step forecast errors.

The study asks three questions. \textbf{RQ1:} What forms of trajectory heterogeneity arise from cross-time ownership? \textbf{RQ2:} How does prediction vary with trajectory partitioning and forecast horizon? \textbf{RQ3:} How do existing algorithms respond to trajectory heterogeneity under the fixed benchmark protocol? We examine these questions across eight disease tasks, ten algorithms, four partitions, and six horizons, connecting the structure of distributed trajectories to forecast quality and learning behavior.

Our contributions follow this evaluation chain. First, Section~\ref{sec:method} defines cross-time ownership, constructs patient-separated clinical trajectories with explicit local-access rules, and specifies a reference world model. Second, Section~\ref{sec:experiments} provides a common evaluation protocol and compares ten federated algorithms across diseases, partitions, and forecast horizons. Third, its coverage and mechanism analyses connect their scores to temporal access, participation and recorded update behavior. Only 7.55\%--21.36\% of pooled-available H32 windows have a locally complete anchor visited in the five-round runs, while FedProx's small aggregate gain is concentrated at shorter horizons. Together, these findings identify what must be measured when evaluating world models trained from distributed trajectories.

%% file: section/related_work.tex
\section{Related Work}
\label{sec:relatedwork}
\subsection{World Models}
World models learn how states evolve under actions from trajectory data, supporting prediction, planning, and policy learning. World Models and PlaNet learn environment dynamics through latent representations, Dreamer learns policies through imagined trajectories, and TD-MPC combines dynamics learning with value estimation for planning~\cite{ha2018worldmodels,hafner2019planet,hafner2020dreamer,hafner2021dreamerv2,hafner2025dreamer,hansen2022tdmpc,hansen2024tdmpc2}. IRIS and DIAMOND extend visual prediction and generation through autoregressive Transformers and diffusion models, respectively~\cite{micheli2023iris,alonso2024diamond}.

Autonomous driving creates practical demands for future-scene prediction, controllable simulation, and action evaluation. GAIA-1 generates driving scenes conditioned on video, text, and actions, while DriveDreamer incorporates structured traffic constraints into diffusion modeling for video generation and action prediction~\cite{hu2023gaia1,wang2024drivedreamer}. Drive-WM connects multiview future generation with driving planning; Vista examines generalization across scenarios, coherent long-horizon rollouts, and multiple forms of action control~\cite{wang2024drivewm,gao2024vista}. DriveWorld uses world models for spatiotemporal pretraining to support perception, prediction, and planning~\cite{min2024driveworld}. NIO's introduction of NIO WorldModel also illustrates automaker interest in this direction~\cite{nio2025worldmodel}.

Disease forecasting, treatment-response modeling, and long-term care similarly require models of evolving patient states. Delphi-2M learns disease progression from longitudinal health records and generates future health trajectories, while Causal Transformer predicts counterfactual outcomes under assumptions that identify time-varying treatment effects~\cite{shmatko2025delphi,melnychuk2022causaltransformer}. EHRWorld and ChronoMedicalWorld study medical world models for long-horizon patient simulation under sequential interventions and action-conditioned kidney-function forecasting, respectively~\cite{mu2026ehrworld,wang2026chronomedicalworld}. EHR-MPC combines generative patient digital twins with model predictive control to study treatment planning using retrospective sepsis data and simulation-based evaluation~\cite{pickard2026ehrmpc}.

Across these applications, trajectory data connect historical states and actions to future changes. Learning these relationships requires diverse, temporally connected experience.

Our study examines world-model learning from distributed trajectory segments. We characterize the histories and future windows available to each client and evaluate their relationship with federated training and multi-step prediction.

\subsection{Federated Learning}
Federated learning combines local training with model aggregation while raw data remain with clients~\cite{mcmahan2017fedavg,kairouz2021advances}. To address client heterogeneity, FedProx and FedDyn regularize local objectives, SCAFFOLD corrects local drift, and FedNova addresses objective inconsistency arising from heterogeneous updates~\cite{li2020fedprox,acar2021feddyn,karimireddy2020scaffold,wang2020fednova}. MOON and FedBN address client differences through representation alignment and local normalization~\cite{li2021moon,li2021fedbn}. Adaptive federated optimization, FedExP, and FedVARP study server adaptation, extrapolation, and update caching; FedCDA, FedLWS, and FedMuon further examine cross-round aggregation, layer-wise shrinking, and alternative optimization mechanisms~\cite{reddi2021adaptive,jhunjhunwala2023fedexp,jhunjhunwala2022fedvarp,wang2024fedcda,shi2025fedlws,takezawa2026fedmuon}. Ditto and pFedHN exploit client-specific information through personalized models~\cite{li2021ditto,shamsian2021pfedhn}.

Practical applications have motivated evaluation resources and collaboration across institutions. LEAF, FedML, and FedScale provide datasets, reference implementations, and system evaluation, while FLamby supplies healthcare tasks with natural institutional partitions~\cite{caldas2018leaf,he2020fedml,lai2022fedscale,terrail2022flamby}. Medical applications include multi-institutional imaging, federated clinical-outcome prediction in EXAM, and multicenter evaluation using electronic health records~\cite{sheller2020medicine,dayan2021exam,dang2022ehr}. These applications establish federated learning as a practical approach to using distributed clinical data.

For temporal data, Fed-TREND studies heterogeneous device time series, continual federated forecasting examines forgetting across seasonal tasks, and FedTDP studies collaborative preparation of trajectory fragments held by regional clients~\cite{yuan2024fedtrend,hallak2025forgetting,zeng2026fedtdp}. Table~\ref{tab:related} compares the coverage of representative world models and federated studies.

Our focus is cross-time learning of action-conditioned world models, where different segments of one trajectory belong to different clients. The benchmark links this temporal ownership to local training windows, client participation, algorithm updates, and multi-step prediction.

\begin{table}[t]
\centering
\caption{Coverage of representative world models and federated studies. Our study jointly evaluates clinical dynamics, federated training, and temporal access within trajectories.}
\label{tab:related}
\begingroup
\fontsize{9}{11}\selectfont
\setlength{\tabcolsep}{5pt}
\renewcommand{\arraystretch}{1.30}
\begin{tabular*}{\textwidth}{@{\extracolsep{\fill}}lcccccc@{}}
\toprule[1.1pt]
& \multicolumn{2}{c}{\textbf{Learning setting}} & \multicolumn{2}{c}{\textbf{Predictive task}} & \multicolumn{2}{c}{\textbf{Trajectory access}}\\
\cmidrule(lr){2-3}\cmidrule(lr){4-5}\cmidrule(l){6-7}
\textbf{Study} & \shortstack{Federated\\training} & \shortstack{Clinical\\tasks} & \shortstack{Multi-step\\prediction} & \shortstack{Action\\conditioning} & \shortstack{Fragmented\\ownership} & \shortstack{Temporal\\access audit}\\
\midrule
Vista~\cite{gao2024vista} & \coverageno & \coverageno & \coverageyes & \coverageyes & \coverageno & \coverageno\\
EHRWorld~\cite{mu2026ehrworld} & \coverageno & \coverageyes & \coverageyes & \coverageyes & \coverageno & \coverageno\\
EHR-MPC~\cite{pickard2026ehrmpc} & \coverageno & \coverageyes & \coverageyes & \coverageyes & \coverageno & \coverageno\\
\midrule
LEAF~\cite{caldas2018leaf} & \coverageyes & \coverageno & \coverageno & \coverageno & \coverageno & \coverageno\\
FedScale~\cite{lai2022fedscale} & \coverageyes & \coverageno & \coverageno & \coverageno & \coverageno & \coverageno\\
FLamby~\cite{terrail2022flamby} & \coverageyes & \coverageyes & \coverageno & \coverageno & \coverageno & \coverageno\\
\midrule
Fed-TREND~\cite{yuan2024fedtrend} & \coverageyes & \coverageno & \coverageyes & \coverageno & \coverageno & \coverageno\\
Continual TSF~\cite{hallak2025forgetting} & \coverageyes & \coverageno & \coverageno & \coverageno & \coverageno & \coverageno\\
FedTDP~\cite{zeng2026fedtdp} & \coverageyes & \coverageno & \coverageno & \coverageno & \coverageyes & \coverageno\\
\midrule
\rowcolor{coverageours}\textbf{Our Study} & \coverageyes & \coverageyes & \coverageyes & \coverageyes & \coverageyes & \coverageyes\\
\bottomrule[1.1pt]
\end{tabular*}\par
\vspace{5pt}
\begin{minipage}{\textwidth}
\fontsize{8}{10}\selectfont
\coverageyes\ explicitly studied; \coverageno\ outside scope or not reported.
Multi-step prediction forecasts a sequence of future states or measurements.
Fragmented ownership assigns segments of one trajectory to different clients.
Temporal access audit quantifies locally available history and future windows.
\end{minipage}
\endgroup
\end{table}

%% file: section/method.tex
\section{Benchmark Construction}
\label{sec:method}
Our benchmark combines a cross-time learning setup, eight clinical trajectory tasks, and a reference world model.

\subsection{Cross-Time Federated Learning Setup}
\label{sec:setting}

\subsubsection{Conditional trajectory prediction}
For a disease cohort, let $S_t\in\mathbb R^F$ denote the physiological state at hourly boundary $t$. Each variable carries its value, validity, freshness, and time since the most recent observation. Let $A_t$ represent the treatment conditions for the interval $[t,t+1)$, and let $C_t$ contain the corresponding context. With a same-stay history of length $L$ and elapsed time $\Delta t_t$, the prediction task is
\begin{equation*}
\widehat S_{t+1}=f_\theta(S_{t-L+1:t},A_t,C_t,\Delta t_t).
\end{equation*}
A latent realization encodes the available history into $z_t$, recursively advances the latent state, and decodes future physiological variables:
\begin{equation*}
\widehat z_{t+r}=F_\theta(\widehat z_{t+r-1},A_{t+r-1},C_{t+r-1},\Delta t_{t+r-1}),\quad r=1,\ldots,h.
\end{equation*}
The model recursively predicts future states along the recorded action and context path.

\subsubsection{Ownership along a trajectory}
Let $c_t$ denote the client owning transition $t$. In cross-time learning, $c_t$ can change within a stay: each transition has one owner, while a patient can contribute different segments to multiple clients.

We assign transitions to severity bands using the latest completed score at the action boundary, then divide each band into ten deterministic patient shards. Three or five bands yield 30 or 50 clients. A patient retains the same shard within a band; worsening or recovery can move subsequent transitions to another band.

Cross-time ownership specifies how a trajectory is divided across clients and which temporal relationships each client can learn. Building on fragmented trajectory preparation~\cite{zeng2026fedtdp}, we define local history and future supervision for action-conditioned prediction.

\subsubsection{Local access and the learning objective}
A client's history contains only indices assigned to that client within the same-stay lookback window. This filtering can retain separated portions of the history when a patient returns to the same client. During local multi-step training, expansion stops when a subsequent transition belongs to another client or leaves the stay. In evaluation, a test anchor is assigned by its starting stage, and the subsequent rollout can cross stage boundaries.

\begin{figure}[t]
\centering
\includegraphics[width=\textwidth]{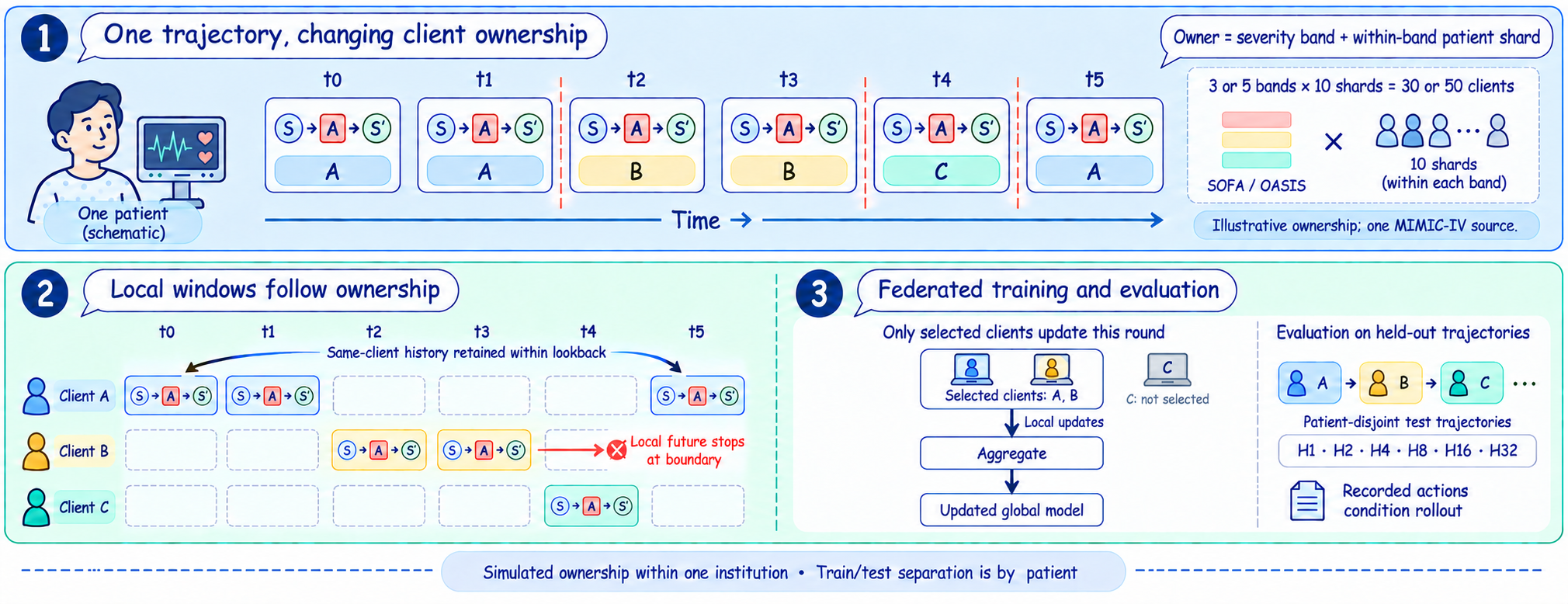}
\caption{Cross-time ownership and local training windows. Clients A, B, and C hold different segments of one stay. History retains same-client positions within the lookback, and future supervision stops at an ownership boundary. Sampled clients update the shared model; evaluation follows patient-disjoint trajectories across stage boundaries.}
\label{fig:setting}
\end{figure}

For client transition sets $\mathcal D_k$, the conventional weighted objective is
\begin{equation*}
\mathcal F_k(\theta)=\frac{1}{|\mathcal D_k|}\sum_{\xi\in\mathcal D_k}\ell(\theta;\xi),\qquad
\mathcal F(\theta)=\sum_k\frac{|\mathcal D_k|}{|\mathcal D|}\mathcal F_k(\theta).
\end{equation*}
Each example $\xi$ uses the history and future targets available under local ownership. The resulting objective depends on both the assigned transitions and their accessible temporal context (Figure~\ref{fig:setting}).

\subsection{Dataset Construction and Client Partitioning}
\label{sec:data}
We construct eight clinical tasks through hourly alignment, patient separation, and severity-based partitioning. Section~\ref{sec:coverage} measures the resulting temporal access.
\subsubsection{Cohorts and temporal cleaning}
We construct the benchmark from MIMIC-IV version 3.1~\cite{johnson2023mimiciv,johnson2024mimiciv31}. Eight disease cohorts are defined by ICD chapter membership. Diagnoses select cohort membership and are not model input features. A patient may belong to multiple cohorts, and each cohort is constructed and evaluated as an independent task. Records and history windows remain within the same ICU stay.

We extract physiological variables from chart, laboratory, output, and related event tables. Field specifications define source items, units, valid ranges, aggregation, and expiry. Measurements are aligned to hourly boundaries by availability time. Within each field's expiry, forward-filled values remain valid; freshness indicates a new measurement, and time since observation records its age. This representation captures irregular measurements on an hourly grid~\cite{che2018grud}.

Treatment inputs combine ongoing exposure at the state boundary with executed events in the next interval, extracted from input, procedure, and medication-administration records. Field specifications distinguish the roles of planned treatments, proxy variables, and device settings. Evaluation supplies the recorded future action path.

\begin{table}[t]
\centering\small\papertablesetup
\caption{Processed cohort sizes. Train and Test count patients within each cohort. State is the number of physiological fields; Exp./Event gives exposure and executed-event dimensions. Cohorts overlap, so totals are membership counts rather than globally unique patients or transitions.}
\label{tab:data}
\begin{adjustbox}{max width=\textwidth}
\begin{tabular}{l||rrrrrr}
\papertoprule
\paperheadrow
\textbf{Cohort} & \textbf{Patients} & \textbf{Train} & \textbf{Test} & \textbf{Transitions} & \textbf{State} & \textbf{Exp./Event}\\
\papergrouprule
Circulatory & 57,069 & 45,655 & 11,414 & 7,439,168 & 27 & 16/54\\
\paperaltrow Digestive & 34,072 & 27,258 & 6,814 & 4,758,459 & 35 & 8/27\\
Endocrine/metabolic & 53,524 & 42,819 & 10,705 & 7,198,590 & 31 & 12/28\\
\paperaltrow Infectious & 23,265 & 18,612 & 4,653 & 4,281,244 & 23 & 9/20\\
Injury/poisoning & 30,765 & 24,612 & 6,153 & 4,553,461 & 35 & 20/41\\
\paperaltrow Neoplasms & 13,914 & 11,131 & 2,783 & 1,602,651 & 28 & 12/30\\
Nervous/sensory & 35,358 & 28,286 & 7,072 & 5,081,808 & 31 & 14/26\\
\paperaltrow Respiratory & 36,174 & 28,939 & 7,235 & 5,956,109 & 20 & 2/15\\
\paperbottomrule
\end{tabular}
\end{adjustbox}
\end{table}

Table~\ref{tab:data} summarizes 41,279,717 state memberships, 40,871,490 transition memberships, and 284,141 patient memberships across the eight cohorts. Disease-specific field dictionaries define the physiological and treatment interfaces.

\subsubsection{Patient separation and normalization}
Within each disease, patients are assigned to Train80 or Test20 with seed 42. Normalization and client construction use Train80. Transitions with an available completed severity score form the eligible population for each score family. Partitions sharing the same eligible population also share normalization statistics.

Future-state supervision and evaluation use valid, fresh targets. Valid historical values provide model inputs, and valid current values support the reconstruction objective in Section~\ref{sec:model}.

\subsubsection{Severity partitions and observed movement}
The main matrix uses SOFA and OASIS scores~\cite{vincent1996sofa,johnson2013oasis}, operationalized as the latest completed 24-hour score at the action boundary. We use these scores to simulate temporal ownership within a single clinical data source. SOFA uses bands $0$--$2$, $3$--$5$, and $\geq6$ for the three-band setting, and $0$, $1$--$2$, $3$--$4$, $5$, and $\geq6$ for five bands. OASIS uses $0$--$25$, $26$--$35$, and $\geq36$, or $0$--$20$, $21$--$25$, $26$--$30$, $31$--$35$, and $\geq36$. For patient $i$ at time $t$, let $b_t\in\{0,\ldots,B-1\}$ be the band determined by the latest completed score, and let $q(i)\in\{0,\ldots,9\}$ be a deterministic patient-shard index. The client index is
\begin{equation*}
k(i,t)=10b_t+q(i).
\end{equation*}
Three or five bands yield 30 or 50 clients. Patient-level train/test separation is fixed before client assignment.

\begin{table}[t]
\centering\small\papertablesetup
\caption{Training-side severity-group statistics for the nervous/sensory cohort before ten-way patient sharding. Multi-client is the percentage of eligible training patients represented in multiple severity groups.}
\label{tab:partition}
\begin{adjustbox}{max width=\textwidth}
\begin{tabular}{l||rrrrr}
\papertoprule
\paperheadrow
\textbf{Partition} & \textbf{Eligible transitions} & \textbf{Min/group} & \textbf{Max/group} & \textbf{Max/min} & \textbf{Multi-client (\%)}\\
\papergrouprule
SOFA-K3 & 4,042,659 & 1,118,953 & 1,542,186 & 1.38 & 68.47\\
\paperaltrow SOFA-K5 & 4,042,659 & 312,139 & 1,230,047 & 3.94 & 86.18\\
OASIS-K3 & 4,071,751 & 1,086,320 & 1,796,094 & 1.65 & 91.68\\
\paperaltrow OASIS-K5 & 4,071,751 & 424,032 & 1,189,337 & 2.80 & 97.52\\
\paperbottomrule
\end{tabular}
\end{adjustbox}
\end{table}

Temporal fragmentation and sample imbalance coexist (Table~\ref{tab:partition}). Under SOFA, 68.47\% and 86.18\% of training patients span multiple groups with three and five bands; under OASIS, these proportions rise to 91.68\% and 97.52\%. The five-band SOFA partition has a 3.94-fold largest-to-smallest group size ratio.

\begin{figure}[p]
\centering
\includegraphics[width=\textwidth]{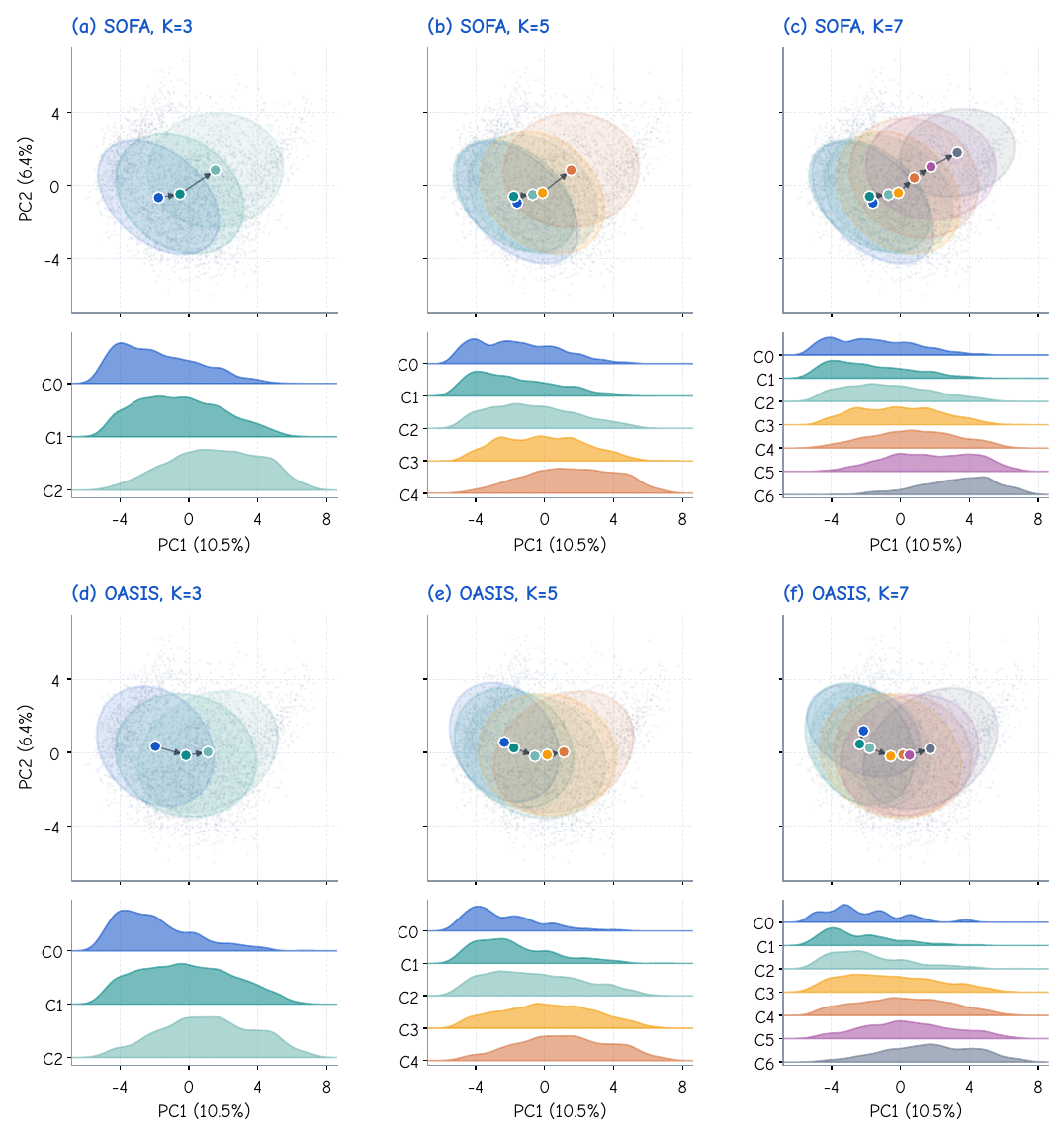}
\caption{SOFA and OASIS partitions in the shared latent space of our model, using the same 4,000 Test20 anchors. Upper panels show points, centroids, and 1.5-standard-deviation covariance ellipses; arrows connect centroids in severity order. Lower panels show smoothed PC1 distributions normalized to their own peaks. PC1 and PC2 explain 10.5\% and 6.4\% of variance. $K$ counts severity groups before patient sharding.}
\label{fig:partition_geometry}
\end{figure}

Figure~\ref{fig:partition_geometry} projects the same 4,000 nervous/sensory Test20 anchors, encoded by our centrally trained model, onto a shared PCA basis. Severity groups overlap while their centroids and PC1 distributions shift, with a more pronounced progression under SOFA. Increasing $K$ reveals finer differences within these distributions. The two displayed components explain 17.0\% of latent variance.

Here $K$ counts severity groups before ten-way sharding. We visualize $K=3,5,7$; the main algorithm experiments use $K=3,5$, corresponding to 30 and 50 clients.

\subsection{Reference World Model Architecture}
\label{sec:model}
\begin{figure}[t]
\centering
\adjincludegraphics[trim={0 {.13\height} 0 0},clip,width=\textwidth]{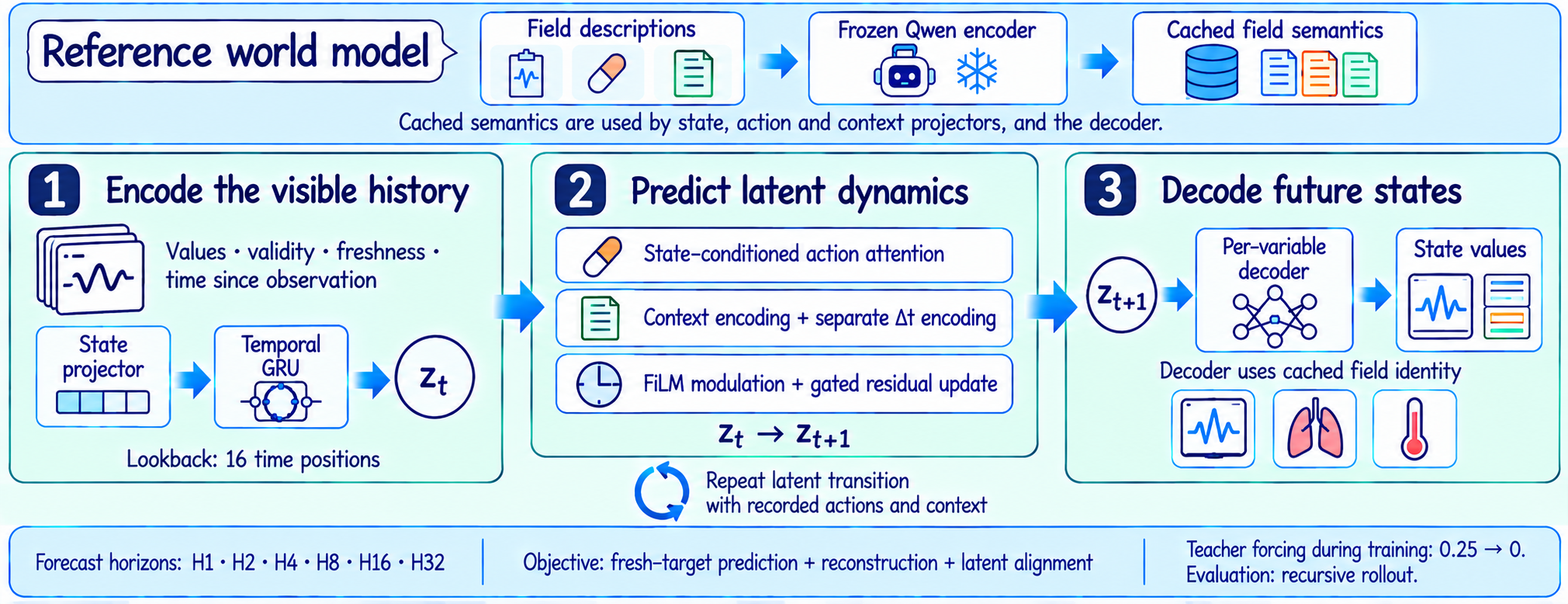}
\caption{Our model encodes visible history, advances an action-conditioned latent state, and decodes physiological values. Cached field semantics support the numerical projectors and decoder. Training combines future prediction, reconstruction, and EMA latent alignment.}
\label{fig:model}
\end{figure}

All algorithms use the same reference architecture to encode visible history, predict conditional latent dynamics, and decode future states.
Figure~\ref{fig:model} organizes the model into history encoding, conditional dynamics, and state decoding. Qwen3.5-4B encodes field descriptions once, offline. The resulting cached vector $e_f$ identifies a field and supplies its semantic representation. All patient-specific computation takes place in the trainable numerical model.

\paragraph{Encode the visible history.}
The state projector combines each field vector $e_f$ with its normalized value, validity, freshness, and time since observation. It pools field tokens into a state representation at each visible position. A temporal gated recurrent unit (GRU) processes these representations within a 16-position lookback and produces $z_t$.

\paragraph{Predict latent dynamics.}
Action tokens combine numerical action inputs with their cached field vectors. A query projected from $z_t$ attends to active action tokens, producing $u_t$; a learned fallback token handles an interval without an active token. Context is pooled and projected into $c_t^{\rm ctx}$, and elapsed time is encoded separately. Their concatenation $q_t=[u_t;c_t^{\rm ctx};d(\log(1+\Delta t_t))]$ conditions feature-wise linear modulation (FiLM) and a gated residual transition:
\begin{align*}
(\gamma_t,\beta_t)&=\operatorname{FiLM}(q_t),&
\widetilde z_t&=(1+\tanh\gamma_t)\odot z_t+\beta_t,\\
r_t&=[\widetilde z_t;q_t],&
z_{t+1}&=\operatorname{LN}\!\left(z_t+g(r_t)\odot v(r_t)\right).
\end{align*}
Here $g$ is a learned sigmoid gate, $v$ predicts the latent increment, $\odot$ denotes elementwise multiplication, and LN denotes layer normalization. This construction lets the encoded state determine action relevance and lets the resulting condition modulate the next transition.

\paragraph{Decode and roll out.}
Each physiological variable has a decoder head that takes the predicted latent state and a projected version of its cached field identity: $\widehat S_{t+h,f}=D_f([z_{t+h};p_D(e_f)])$. The model predicts numerical state values. At evaluation, it encodes the observed initial history once and repeatedly applies the latent transition, with recorded actions and context supplied at each step. Future observed states are used for scoring. Our model uses a 384-dimensional latent state, state-token and action dimensions of 192, context dimension 96, and dynamics/decoder hidden dimensions 768/384.

\paragraph{Train with locally available targets.}
The training objective is $\mathcal L=\mathcal L_{\rm future}+0.2\mathcal L_{\rm recon}+0.1\mathcal L_{\rm latent}$. The future term averages feature-balanced, normalized L1 losses over available supervised horizons $\{1,2,4,8,16,32\}$, using valid and fresh targets. Reconstruction decodes the initial latent state and compares it with valid current values. Latent alignment compares predicted latents with target-history representations from an exponential-moving-average (EMA) encoder, using decay 0.995. The EMA encoder starts from the global online encoder each round and remains local. During training, teacher forcing replaces a predicted latent with the EMA target-history latent at the scheduled probability, which decreases from 0.25 to zero. Evaluation uses recursive prediction throughout. Ownership boundaries restrict the future steps available to these objectives.

An earlier architecture comparison is reported in the experiments using its original model configuration and metric.

%% file: section/experiments.tex
\section{Experiments}
\label{sec:experiments}
Our experiments examine trajectory heterogeneity through three linked questions: what changes when trajectories are divided across clients (RQ1), how prediction varies with partition granularity and forecast horizon (RQ2), and how algorithm updates respond to those segments (RQ3). SOFA/OASIS and three/five severity bands vary the temporal partition; H1--H32 expose its consequences at different prediction lengths.

\subsection{Experimental setup}
\label{sec:experimentalsetup}
\subsubsection{Algorithms and participation}

The main matrix evaluates FedAvg~\cite{mcmahan2017fedavg}, FedProx~\cite{li2020fedprox}, server-momentum FedAvgM, FedAdam~\cite{reddi2021adaptive}, SCAFFOLD~\cite{karimireddy2020scaffold}, FedExP~\cite{jhunjhunwala2023fedexp}, FedVARP~\cite{jhunjhunwala2022fedvarp}, FedCDA~\cite{wang2024fedcda}, FedLWS~\cite{shi2025fedlws}, and FedMuon~\cite{takezawa2026fedmuon}. These methods cover proximal local objectives, gradient correction, server adaptation, extrapolation, cached updates, cross-round model selection, weight shrinking, and local optimization changes.

Experiments cover eight diseases and four partitions: SOFA-30, SOFA-50, OASIS-30, and OASIS-50. Each run uses five communication rounds and one local epoch per participation. Each round samples one shard per severity band, giving 10\% client participation. Seed 42 fixes the schedule, with clients able to recur across rounds. FedAvg weights selected models by their training-anchor counts. We report round-5 performance.

Local update rules differ. FedAvg uses AdamW and SCAFFOLD uses SGD, both at a nominal learning rate of $10^{-4}$; FedMuon uses its LMO-based optimizer. FedProx uses $\mu=0.01$, FedAvgM server momentum 0.9, and FedAdam server learning rate 0.01, moments 0.9/0.99, and stabilizer 0.001. All methods share the model architecture, data partition, and participation schedule.

\subsubsection{Scoring and comparison rules}
We report Fresh Feature--Patient Macro normalized MAE (nMAE). For patient $i$, feature $f$, and horizon $h$, let $\mathcal T_{if}^{(h)}$ contain eligible test anchors whose target at $t+h$ is valid and fresh. With training standard deviation $\sigma_f^{\rm train}$, the score is
\begin{align}
e_{if}^{(h)}&=\frac{1}{|\mathcal T_{if}^{(h)}|}
\sum_{t\in\mathcal T_{if}^{(h)}}
\frac{|\widehat S_{i,t+h,f}^{(t)}-S_{i,t+h,f}|}{\sigma_f^{\rm train}},\notag\\
\operatorname{nMAE}_h&=\frac{1}{|\mathcal F_h|}\sum_{f\in\mathcal F_h}
\frac{1}{|\mathcal P_{f,h}|}\sum_{i\in\mathcal P_{f,h}}e_{if}^{(h)}.
\label{eq:nmae}
\end{align}
Here $\widehat S_{i,t+h,f}^{(t)}$ is the prediction issued at anchor $t$, $\mathcal P_{f,h}$ contains patients with at least one scored target for feature $f$, and $\mathcal F_h$ contains features with scored patients. Equation~\ref{eq:nmae} averages time points within each patient--feature pair, patients within each feature, and finally features. We evaluate the full eligible Test20 population at $\mathcal H=\{1,2,4,8,16,32\}$ and report $\operatorname{nMAE}=|\mathcal H|^{-1}\sum_{h\in\mathcal H}\operatorname{nMAE}_h$. Cross-disease summaries weight diseases equally.

We also report normalized MSE (nMSE) with the same valid, fresh targets and feature--patient weighting:
\begin{equation}
\operatorname{nMSE}_h=\frac{1}{|\mathcal F_h|}\sum_{f\in\mathcal F_h}
\frac{1}{|\mathcal P_{f,h}|}\sum_{i\in\mathcal P_{f,h}}
\frac{1}{|\mathcal T_{if}^{(h)}|}\sum_{t\in\mathcal T_{if}^{(h)}}
\left(\frac{\widehat S_{i,t+h,f}^{(t)}-S_{i,t+h,f}}{\sigma_f^{\rm train}}\right)^2.
\label{eq:nmse}
\end{equation}
Equation~\ref{eq:nmse} gives greater weight to large prediction errors; its overall score also averages the six horizons.

Algorithm gains are computed relative to FedAvg within each disease--configuration pair and then averaged. The centralized references use 10\% of eligible training data for five epochs.

\subsubsection{Reference-model comparison}
\begin{figure}[!tbp]
\centering
\begin{minipage}[t]{.49\textwidth}
\vspace{0pt}\captionsetup{font=footnotesize}
\centering\footnotesize\papertablesetup\setlength{\tabcolsep}{3pt}
\captionof{table}{Earlier centralized architecture comparison using Fresh patient-macro nMAE.}
\label{tab:models}
\begin{adjustbox}{max width=\linewidth}
\begin{tabular}{l||rrr}
\papertoprule
\paperheadrow
\textbf{Cohort} & \shortstack{\textbf{Best}\\\textbf{structured}} & \shortstack{\textbf{Semantic}\\\textbf{model}} & \shortstack{\textbf{Gain}\\\textbf{(\%)}}\\
\papergrouprule
Circulatory & 0.489639 & 0.463037 & 5.43\\
\paperaltrow Digestive & 0.499552 & 0.444876 & 10.94\\
Endocrine/metabolic & 0.523904 & 0.474186 & 9.49\\
\paperaltrow Infectious & 0.503779 & 0.478182 & 5.08\\
Injury/poisoning & 0.532528 & 0.478505 & 10.14\\
\paperaltrow Neoplasms & 0.555474 & 0.518035 & 6.74\\
Nervous/sensory & 0.518287 & 0.477117 & 7.94\\
\paperaltrow Respiratory & 0.507782 & 0.494651 & 2.59\\
\paperbottomrule
\end{tabular}
\end{adjustbox}
\end{minipage}\hfill
\begin{minipage}[t]{.48\textwidth}
\vspace{0pt}\captionsetup{font=footnotesize}
\centering
\includegraphics[width=\linewidth]{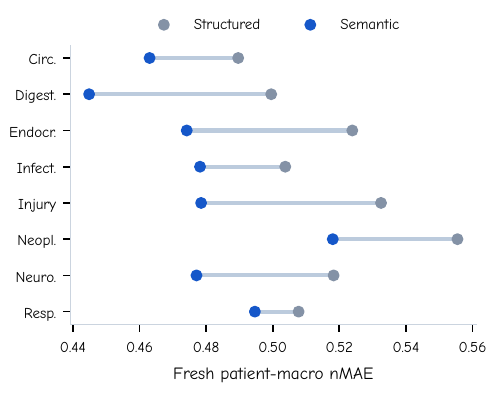}
\caption{Paired scores for the best structured model and the semantic model.}
\label{fig:architecture_comparison}
\end{minipage}
\end{figure}
The semantic latent model achieves lower overall error than the best structured candidate in all eight cohorts (Table~\ref{tab:models}). This earlier architecture comparison uses the original model configuration and patient-macro metric; the federated experiments use our current model and feature--patient metric.

Figure~\ref{fig:architecture_comparison} displays the paired architecture results for each disease.

\subsection{Characterizing trajectory heterogeneity (RQ1)}
\label{sec:coverage}
\textbf{RQ1: What forms of trajectory heterogeneity arise from cross-time ownership?} Severity groups occupy shifted regions of the shared latent space (Figure~\ref{fig:partition_geometry}). We examine their temporal structure by comparing local and pooled access under SOFA/OASIS with 30/50 clients, then measuring the windows reached by sampled participation. We distinguish three forms: shifts in state distributions, unequal temporal context, and changes in which severity groups supply long-window examples.
\subsubsection{Temporal context under local ownership}

\label{sec:accessaudit}
We measure temporal access for every eligible training anchor across the 32 disease--partition settings. Pooled access uses the eligible training union; local access uses the owning client. Both follow stay boundaries and score eligibility.

Complete-window retention at horizon $h$ is the fraction of pooled-available $h$-step windows whose transitions remain in the starting client. History retention uses the model's 16-position lookback and includes earlier same-client positions separated by another client's segment.

\begin{table}[!tbp]
\centering\small\papertablesetup
\caption{Temporal access on Train80, averaged equally across eight diseases. History local/pooled is the mean number of visible positions in a 16-position lookback. History loss is the proportion of eligible anchors that lose at least one history position solely due to client ownership. H8--H32 columns report complete-window retention conditional on pooled availability. The last column gives the minimum and maximum H32 retention across diseases.}
\label{tab:access}
\begin{adjustbox}{max width=\textwidth}
\begin{tabular}{l||rrrrrr}
\papertoprule
\paperheadrow
\textbf{Partition} & \textbf{History local/pooled} & \textbf{History loss (\%)} & \textbf{H8 (\%)} & \textbf{H16 (\%)} & \textbf{H32 (\%)} & \textbf{H32 range (\%)}\\
\papergrouprule
SOFA-30 & 12.94 / 14.80 & 28.37 & 85.40 & 72.13 & 53.34 & 52.36--54.56\\
\paperaltrow SOFA-50 & 12.14 / 14.80 & 39.08 & 79.31 & 62.59 & 42.17 & 40.55--44.09\\
OASIS-30 & 11.89 / 14.81 & 45.64 & 74.96 & 57.41 & 35.88 & 34.72--37.04\\
\paperaltrow OASIS-50 & 10.44 / 14.81 & 61.34 & 62.22 & 40.95 & 21.71 & 19.95--23.84\\
\paperbottomrule
\end{tabular}
\end{adjustbox}
\end{table}

Table~\ref{tab:access} shows a systematic loss of temporal context. Pooled access provides approximately 14.8 visible history positions on average. Local visibility reduces this to 12.94 and 12.14 under SOFA-30 and SOFA-50, and to 11.89 and 10.44 under OASIS-30 and OASIS-50. Between 28.37\% and 61.34\% of anchors lose at least one history position relative to the pooled eligible union.

\begin{figure}[!tbp]
\centering
\includegraphics[width=\textwidth]{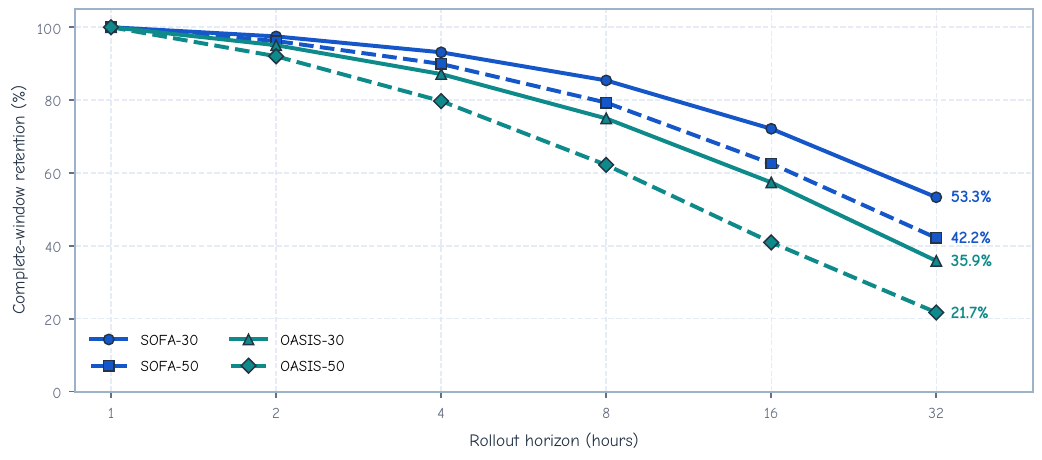}
\caption{Retention of complete local training windows relative to pooled access. Curves average per-disease retention ratios, using the same eligible anchors within each comparison. H1 is fully retained, while longer horizons increasingly encounter ownership changes.}
\label{fig:access}
\end{figure}

Future-window availability decreases more sharply with horizon (Figure~\ref{fig:access}). All eligible anchors retain H1, but the disease-macro H32 retention falls to 53.34\% for SOFA-30, 42.17\% for SOFA-50, 35.88\% for OASIS-30, and 21.71\% for OASIS-50. The pattern holds across all eight cohorts: OASIS-50 retains 19.95\%--23.84\% of the pooled-available H32 windows, compared with 52.36\%--54.56\% for SOFA-30. Local temporal continuity therefore determines how much of the transition pool supports long-window learning.

\subsubsection{Long-window support changes the severity composition}
Window loss varies substantially across severity groups. Let $N_g$ count eligible anchors in group $g$ and $L_{g,h}$ count its locally complete $h$-step windows. Their shares are
\begin{equation}
p_g=\frac{N_g}{\sum_jN_j},\qquad
q_{g,h}=\frac{L_{g,h}}{\sum_jL_{j,h}}.
\label{eq:composition}
\end{equation}
The change from $p_g$ to $q_{g,h}$ measures how local temporal continuity reshapes training support. We pool the ten shards within each severity group, calculate shares within each disease, and then average across diseases.

\begin{figure}[!tbp]
\centering
\includegraphics[width=\textwidth]{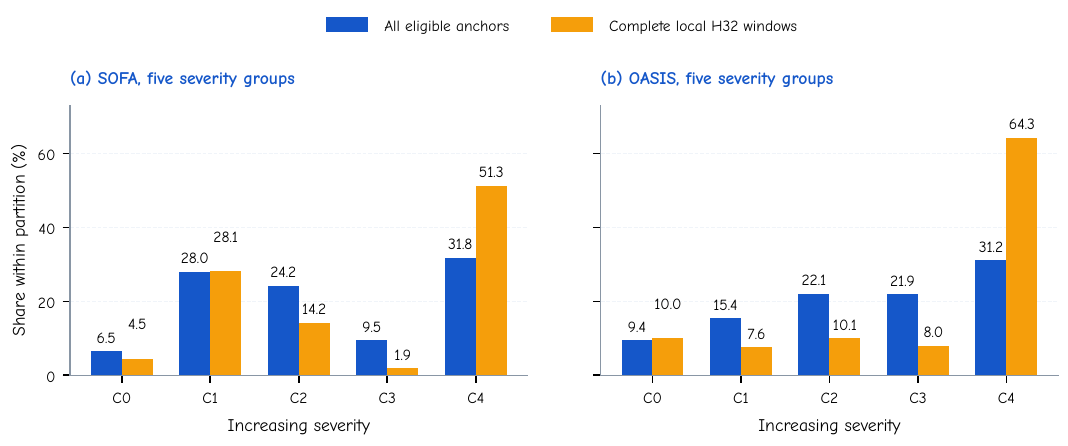}
\caption{Temporal continuity changes the severity composition of training examples. Bars compare each group's share of eligible anchors ($p_g$) with its share of complete local H32 windows ($q_{g,32}$) in five-band partitions. Shares are calculated within each disease before averaging across eight diseases. Groups aggregate ten patient shards; window counts precede participation sampling.}
\label{fig:composition}
\end{figure}

In the five-band OASIS partition, the highest-severity group supplies 31.16\% of all anchors but 64.26\% of complete local H32 windows (Figure~\ref{fig:composition}). The 31--35 group supplies 21.94\% of anchors but only 8.03\% of H32 windows; its H32 retention is 7.60\%, compared with 39.73\% in the highest group. SOFA shows a similar concentration: its highest group rises from 31.82\% of anchors to 51.31\% of H32 windows, while the score-5 group falls from 9.49\% to 1.89\%.

This is horizon-dependent composition shift. An anchor supports H1 whenever its starting transition is eligible, whereas a complete H32 example requires subsequent transitions to remain under the same ownership. Narrow intermediate bands lose windows at either boundary as severity rises or falls. Consequently, long-window learning concentrates on groups that retain continuous segments. A client's anchor count therefore captures a different training distribution from its long-window support. The shared latent overlap in Figure~\ref{fig:partition_geometry} can coexist with a much larger difference in accessible temporal relationships.

\subsubsection{Coverage after sampled participation}

For horizon $h$, let $P_h$ count pooled-available complete-window anchors, $L_h$ the subset that remains complete under local ownership, and $V_h$ the subset of $L_h$ whose owning client participates at least once. All counts use the same eligible training union. The identity
\begin{equation}
\frac{V_h}{P_h}=\frac{L_h}{P_h}\,\frac{V_h}{L_h}
\label{eq:coverage}
\end{equation}
separates ownership retention from participation coverage among locally complete windows. The three disjoint categories $V_h$, $L_h-V_h$, and $P_h-L_h$ also distinguish visited complete windows, unvisited locally complete windows, and windows truncated by ownership. Ratios are calculated within each configuration before averaging across diseases.

\begin{table}[!tbp]
\centering\small\papertablesetup
\caption{Recorded five-round coverage, averaged across eight diseases per partition. Anchors seen counts distinct visited training anchors; anchor passes counts visits including repeats divided by the eligible-anchor total. H32 seen is $V_{32}/P_{32}$ from Equation~\ref{eq:coverage}.}
\label{tab:coverage}
\begin{adjustbox}{max width=\textwidth}
\begin{tabular}{l||rrrr}
\papertoprule
\paperheadrow
\textbf{Partition} & \textbf{Clients seen} & \textbf{Anchors seen (\%)} & \textbf{Anchor passes} & \textbf{H32 seen (\%)}\\
\papergrouprule
SOFA-30 & 12/30 & 39.66 & 0.504 & 21.36\\
\paperaltrow SOFA-50 & 20/50 & 38.62 & 0.505 & 15.76\\
OASIS-30 & 12/30 & 40.64 & 0.504 & 14.80\\
\paperaltrow OASIS-50 & 20/50 & 38.83 & 0.506 & 7.55\\
\paperbottomrule
\end{tabular}
\end{adjustbox}
\end{table}

\begin{figure}[!tbp]
\centering
\includegraphics[width=\textwidth]{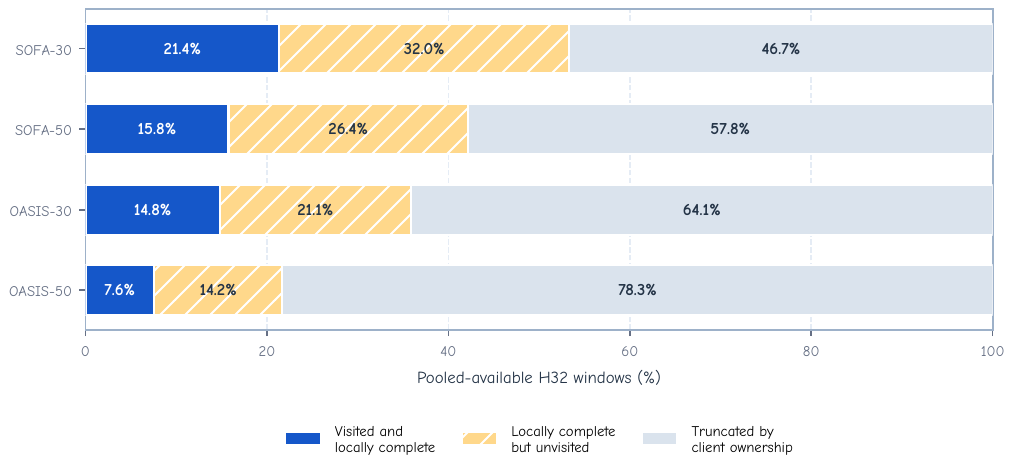}
\caption{Only 7.55\%--21.36\% of pooled-available H32 windows have a locally complete anchor visited during training. The stacked categories share one denominator and exhaust the pooled-available windows. Ownership changes and nonparticipation reduce complete-window coverage through separate paths.}
\label{fig:coverage}
\end{figure}

All schedules visit 12 of 30 or 20 of 50 clients. The exact anchor coverage differs with client sizes: 38.62\%--40.64\% of eligible anchors are visited, while repeated participation gives approximately 0.504--0.506 anchor passes (Table~\ref{tab:coverage}). The decomposition in Figure~\ref{fig:coverage} combines these logs with the access audit. It gives H32 coverage of 21.36\%, 15.76\%, 14.80\%, and 7.55\% for SOFA-30, SOFA-50, OASIS-30, and OASIS-50, respectively.

\noindent\textbf{Finding 1: trajectory heterogeneity changes both the amount and the composition of temporal supervision.} State distributions overlap, but local context lengths and long-window support differ strongly across groups. In OASIS-50, the highest-severity group supplies 31.16\% of anchors and 64.26\% of complete H32 windows; sampled participation further reduces overall visited H32 coverage to 7.55\%.

\subsection{Prediction under trajectory heterogeneity (RQ2)}
\label{sec:results}
\textbf{RQ2: How does prediction vary with trajectory partitioning and forecast horizon?} We first pair three-band and five-band partitions within each disease and score family, then compare ten algorithms across the four partitions and six horizons. This tests the sensitivity of forecasting to partition granularity and whether algorithm gains extend to longer trajectories.
\subsubsection{Finer severity partitions and forecasting error}
We pair three-band and five-band configurations within each disease and score family. Each pair has the same eligible-anchor count, pooled H32-window count, pooled mean history length, and centralized reference. Moving from 30 to 50 clients reduces mean H32 retention by 11.17 percentage points under SOFA and 14.17 points under OASIS (Table~\ref{tab:granularity}). FedAvg nMAE increases in 15 of the 16 pairs. Averaging the relative changes within each score family gives increases of 1.24\% and 3.27\%, respectively; the sole decrease is 0.11\% for the circulatory SOFA cohort.

\begin{table}[!tbp]
\centering\small\papertablesetup
\caption{Paired effects of changing from three to five severity bands. Retention changes are percentage points. nMAE changes are relative to the three-band FedAvg score at each horizon; Mean uses the six-horizon score. All changes are calculated per disease and then averaged.}
\label{tab:granularity}
\begin{adjustbox}{max width=\textwidth}
\begin{tabular}{l||rrrrrr}
\papertoprule
\paperheadrow
\textbf{Score} & \textbf{$\Delta$ H32 retention (pp)} & \textbf{H1 (\%)} & \textbf{H8 (\%)} & \textbf{H32 (\%)} & \textbf{Mean (\%)} & \textbf{Error increases}\\
\papergrouprule
SOFA & -11.17 & +1.70 & +1.04 & +0.73 & +1.24 & 7/8\\
\paperaltrow OASIS & -14.17 & +4.26 & +2.90 & +2.11 & +3.27 & 8/8\\
\paperbottomrule
\end{tabular}
\end{adjustbox}
\end{table}
Finer severity partitions combine lower temporal retention with higher forecasting error. They distribute the same eligible data among more clients and increase the selected clients per round from three to five at the fixed 10\% participation rate.

\paragraph{The partition penalty is already present at H1.}
The relative penalty decreases from 1.70\% at H1 to 0.73\% at H32 under SOFA, and from 4.26\% to 2.11\% under OASIS. This pattern also holds in absolute error, separating it from changes in the relative-error denominator. Let $E_h^{(K)}$ denote FedAvg nMAE with $K$ severity bands and $\Delta_h=E_h^{(5)}-E_h^{(3)}$. Then
\begin{equation*}
\Delta_{32}=\Delta_1+
\left[(E_{32}^{(5)}-E_1^{(5)})-(E_{32}^{(3)}-E_1^{(3)})\right].
\end{equation*}
The mean absolute penalty falls from 0.00990 to 0.00448 under SOFA and from 0.02347 to 0.01251 under OASIS; $\Delta_{32}<\Delta_1$ in 15 of 16 pairs. Thus, much of the observed separation is established at the first predicted step. H1 windows remain available, while mean local history falls from 12.94 to 12.14 positions under SOFA and from 11.89 to 10.44 under OASIS. The partition changes both historical context and client-specific state distributions. The error gap is therefore visible even at H1, where future-window retention is complete: access to the immediate target coexists with differences in historical context and state composition.

\begin{figure}[!tbp]
\centering
\includegraphics[width=.87\textwidth]{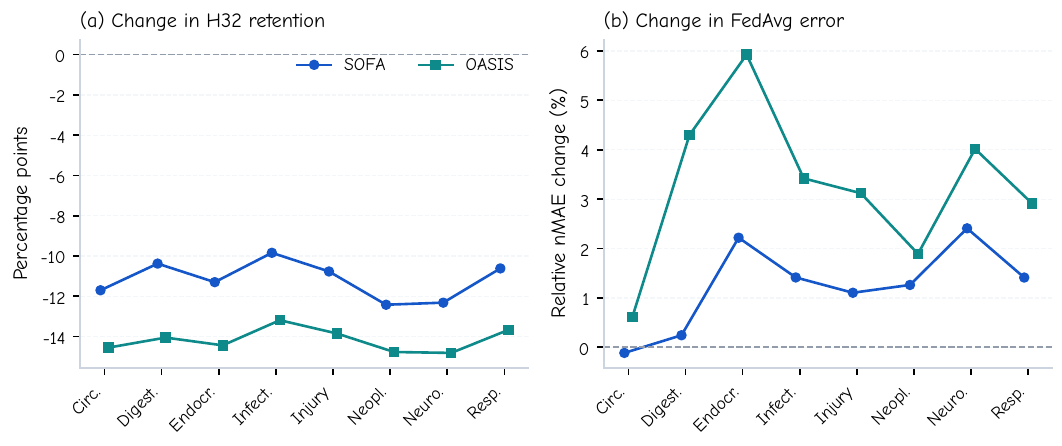}
\caption{Disease-wise changes from three to five severity bands. Left: complete H32-window retention change in percentage points. Right: relative change in the six-horizon FedAvg nMAE. Colors identify SOFA and OASIS.}
\label{fig:granularity_by_disease}
\end{figure}

Figure~\ref{fig:granularity_by_disease} resolves the paired effects by disease: all cohorts lose complete-window support, while prediction sensitivity varies across cohorts.

\subsubsection{Existing federated methods under the benchmark protocol}
\begin{figure}[!tbp]
\centering
\includegraphics[width=\textwidth]{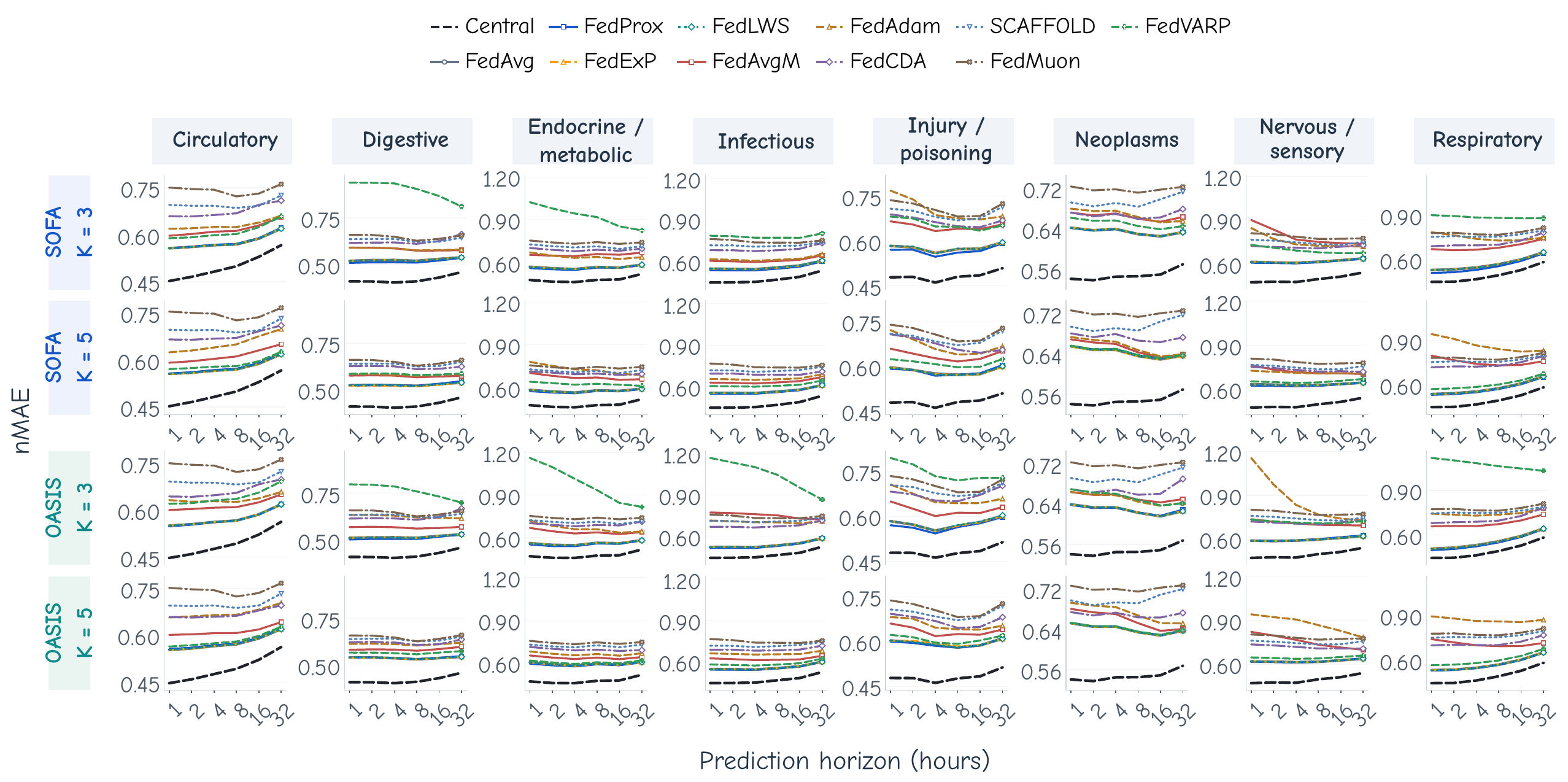}
\caption{nMAE ($\downarrow$) across eight disease cohorts (columns) and four trajectory partitions (rows). Curves show Central and ten federated methods at H1--H32. Vertical limits are shared within each disease column.}
\label{fig:algorithm_matrix}
\end{figure}

\begin{figure}[!tbp]
\centering
\includegraphics[width=\textwidth]{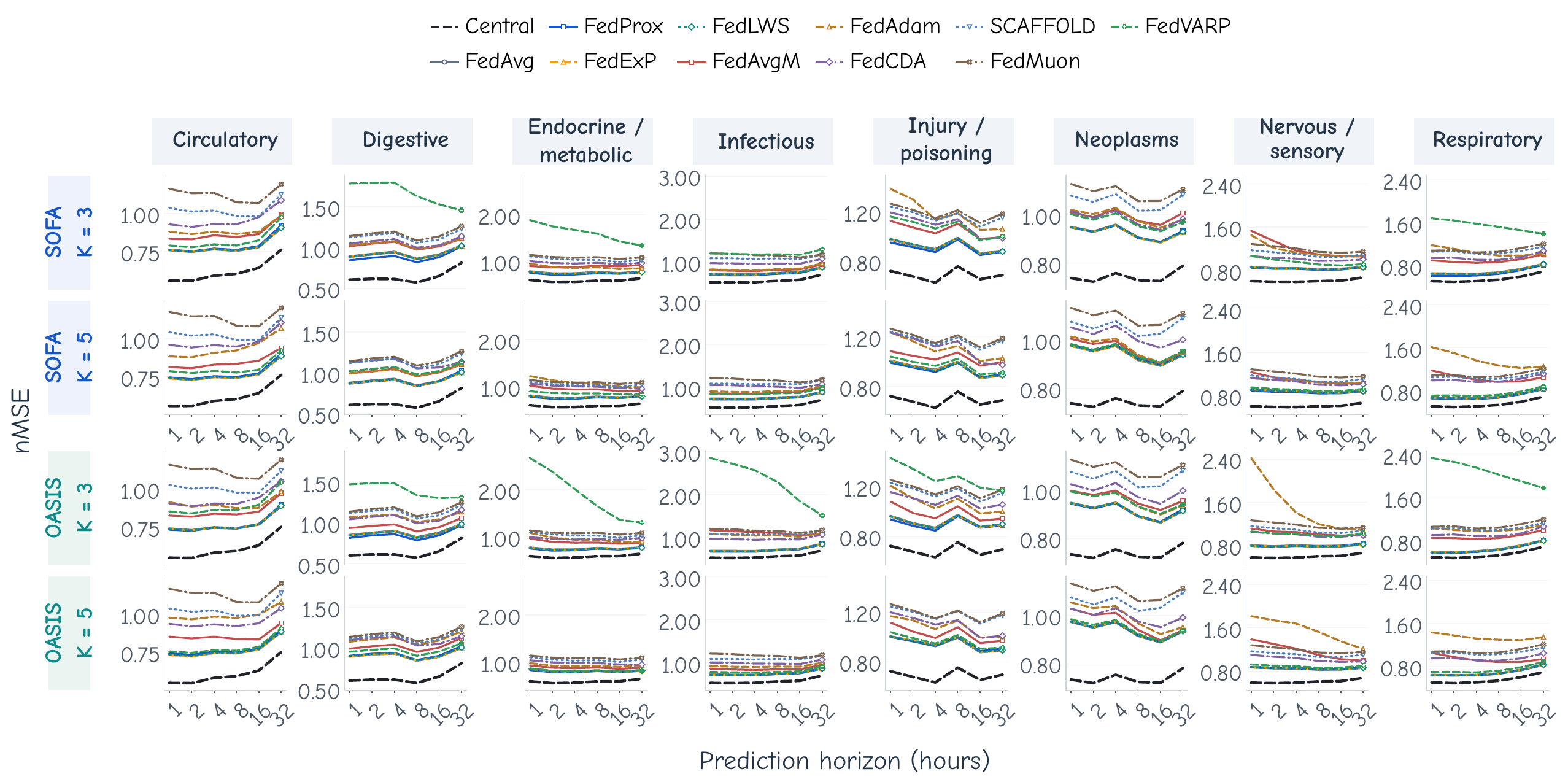}
\caption{nMSE ($\downarrow$) for the same diseases, partitions, methods, and horizons as Figure~\ref{fig:algorithm_matrix}. Errors are normalized by training feature variance; vertical limits are shared within each disease column.}
\label{fig:algorithm_matrix_nmse}
\end{figure}

Figures~\ref{fig:algorithm_matrix} and~\ref{fig:algorithm_matrix_nmse} present all eight diseases as columns and the four trajectory partitions as rows. Each panel shows absolute nMAE or nMSE from H1 to H32 for Central and the ten federated methods. Central uses the corresponding disease and score-family reference, shared between the three-band and five-band panels. Reading down a column compares partitions within the same disease; reading across a row compares disease-specific horizon profiles.

Figure~\ref{fig:algorithm_matrix} shows three distinct patterns. FedProx gives small improvements, averaging 0.56\% across paired configurations. FedExP and FedLWS stay close to FedAvg, with average changes of approximately $-0.04\%$ each. Several other methods increase error across much of the matrix: the mean paired changes are $+13.04\%$ for FedAvgM, $+19.24\%$ for FedAdam, $+22.21\%$ for SCAFFOLD, and $+26.78\%$ for FedMuon. The nMSE matrix shows the same broad ordering: FedProx reduces mean paired error by 0.90\%, while FedAvgM, FedAdam, SCAFFOLD, and FedMuon increase it by 18.44\%, 30.38\%, 32.70\%, and 39.62\%, respectively. These methods perform worse under both absolute error and the squared-error score, which gives greater weight to large prediction errors.

\begin{figure}[!tbp]
\centering
\begin{minipage}[t]{.49\textwidth}
\vspace{0pt}\captionsetup{font=footnotesize}
\centering\footnotesize\papertablesetup\setlength{\tabcolsep}{3pt}
\captionof{table}{Consistency across 32 disease--partition settings, grouped by paired error changes relative to FedAvg.}
\label{tab:gains}
\begin{adjustbox}{max width=\linewidth}
\begin{tabular}{l||rrr}
\papertoprule
\paperheadrow
\textbf{Method} & \shortstack{\textbf{Improve}\\$>1\%$} & \shortstack{\textbf{Within}\\$\pm1\%$} & \shortstack{\textbf{Worsen}\\$>1\%$}\\
\papergrouprule
FedProx & 7 & 25 & 0\\
\paperaltrow FedExP & 0 & 32 & 0\\
FedLWS & 0 & 32 & 0\\
\paperaltrow FedAvgM & 0 & 0 & 32\\
FedAdam & 0 & 0 & 32\\
\paperaltrow FedCDA & 0 & 0 & 32\\
SCAFFOLD & 0 & 0 & 32\\
\paperaltrow FedMuon & 0 & 0 & 32\\
FedVARP & 0 & 2 & 30\\
\paperbottomrule
\end{tabular}
\end{adjustbox}
\end{minipage}\hfill
\begin{minipage}[t]{.48\textwidth}
\vspace{0pt}\captionsetup{font=footnotesize}
\centering
\includegraphics[width=\linewidth]{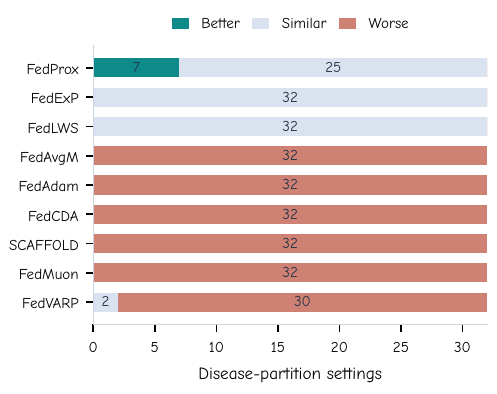}
\caption{Consistency across 32 settings. Better/worse denote changes beyond $\pm1\%$ relative to FedAvg; similar denotes changes within this range.}
\label{fig:algorithm_consistency}
\end{minipage}
\end{figure}

Table~\ref{tab:gains} separates average performance from consistency. FedProx improves by more than 1\% in seven configurations, while its other 25 changes lie within $\pm1\%$. Every FedExP and FedLWS result is within that range. FedVARP is more variable: its paired degradation ranges from 0.16\% to 93.16\%, and its OASIS-30 macro score is substantially worse than its OASIS-50 score. This variation motivates examining how participation history interacts with cached updates.

Figure~\ref{fig:algorithm_consistency} visualizes how consistently each method improves or worsens across the benchmark settings.

\subsubsection{Rollout horizon and centralized references}
\begin{figure}[!tbp]
\centering
\includegraphics[width=\textwidth]{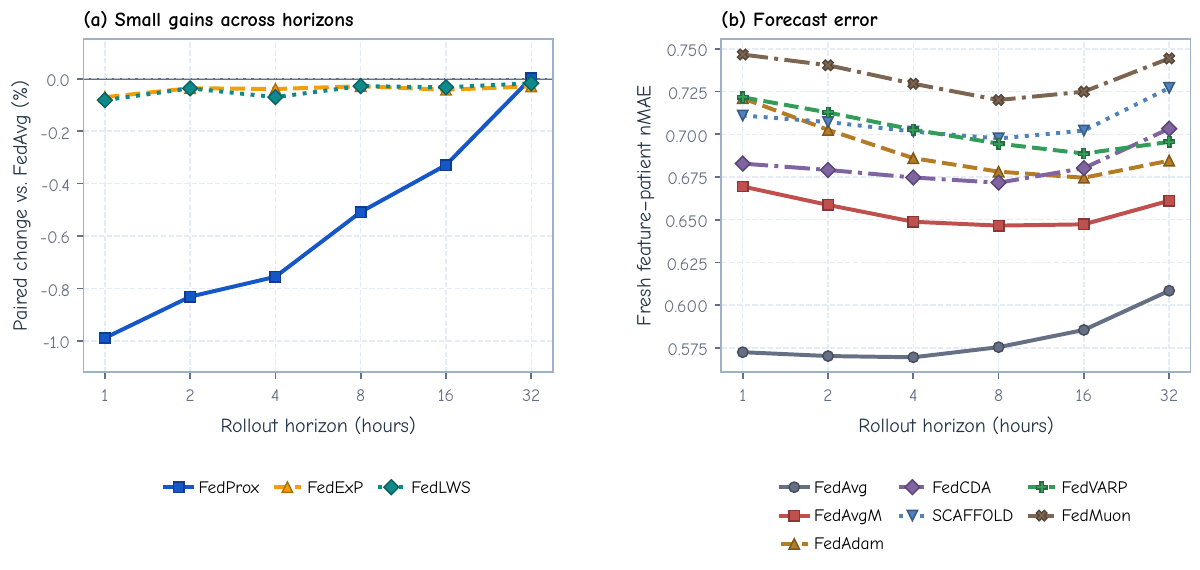}
\caption{Horizon profiles averaged equally over eight diseases and four configurations. The left panel shows mean paired relative changes for methods close to FedAvg (negative is better); the right shows absolute macro nMAE for the remaining methods and FedAvg. Both use all 32 settings and six horizons.}
\label{fig:horizon}
\end{figure}
Figure~\ref{fig:horizon} shows the rollout profiles, while Table~\ref{tab:pairedh} computes each relative change within a matched disease--partition pair before averaging. FedProx's mean paired gain falls from 0.989\% at H1 to an effectively zero change at H32. Fourteen H1 settings improve by more than 1\%, compared with one at H32. At H32, exactly 16 settings improve and 16 worsen. Leaving out any one disease keeps the H1 mean improvement between 0.874\% and 1.136\%; the corresponding H32 mean change ranges from $-0.045\%$ to $+0.092\%$. Thus, the short-horizon benefit and near-zero long-horizon effect are stable to removing one cohort. Together with the granularity comparison, this separates two requirements: fitting transitions from locally visible histories and preserving predictive gains through recursive rollout. FedProx improves the former score, while its advantage disappears over longer trajectories.

\begin{table}[!tbp]
\centering\small\papertablesetup
\caption{FedProx's gain concentrates at shorter horizons. Relative nMAE changes are computed against FedAvg in each of 32 settings, then averaged; negative is better. The 1\% count is an effect-size summary.}
\label{tab:pairedh}
\begin{adjustbox}{max width=\textwidth}
\begin{tabular}{l||rrrrrr}
\papertoprule
\paperheadrow
\textbf{FedProx vs. FedAvg} & \textbf{H1} & \textbf{H2} & \textbf{H4} & \textbf{H8} & \textbf{H16} & \textbf{H32}\\
\papergrouprule
Mean change (\%) & -0.989 & -0.831 & -0.755 & -0.508 & -0.329 & +0.003\\
\paperaltrow Improvement $>1\%$ & 14/32 & 15/32 & 12/32 & 6/32 & 5/32 & 1/32\\
\paperbottomrule
\end{tabular}
\end{adjustbox}
\end{table}

\begin{figure}[!tbp]
\centering
\includegraphics[width=\textwidth]{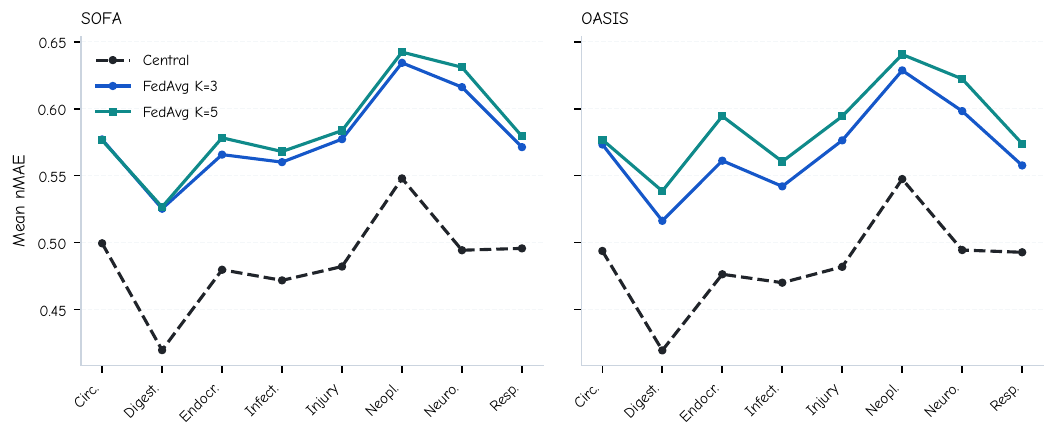}
\caption{Central and FedAvg across all eight diseases and both score families. Points average nMAE over six horizons. The reference and federated configurations follow Section~\ref{sec:experimentalsetup}.}
\label{fig:central_by_disease}
\end{figure}

\begin{table}[!tbp]
\centering\small\papertablesetup
\caption{Centralized and FedAvg nMAE by disease under SOFA-30, using the reference specified in Section~\ref{sec:experimentalsetup}.}
\label{tab:central}
\begin{adjustbox}{max width=\textwidth}
\begin{tabular}{l||rrr}
\papertoprule
\paperheadrow
\textbf{Cohort} & \textbf{Central-10\%} & \textbf{FedAvg} & \textbf{Gap (\%)}\\
\papergrouprule
Circulatory & 0.4995 & 0.5770 & +15.53\\
\paperaltrow Digestive & 0.4197 & 0.5253 & +25.16\\
Endocrine/metabolic & 0.4797 & 0.5657 & +17.93\\
\paperaltrow Infectious & 0.4718 & 0.5601 & +18.71\\
Injury/poisoning & 0.4821 & 0.5773 & +19.73\\
\paperaltrow Neoplasms & 0.5479 & 0.6342 & +15.75\\
Nervous/sensory & 0.4943 & 0.6161 & +24.64\\
\paperaltrow Respiratory & 0.4957 & 0.5713 & +15.26\\
\paperbottomrule
\end{tabular}
\end{adjustbox}
\end{table}
Figure~\ref{fig:central_by_disease} extends the SOFA-30 comparison in Table~\ref{tab:central} to both score families and both partition granularities. Section~\ref{sec:diagnosis} examines algorithm behavior within this benchmark.

\noindent\textbf{Finding 2: trajectory partitioning affects short-step fit and long-horizon learning differently.} Finer partitions raise error in 15 of 16 pairs, but their error penalty is largest at short horizons. FedProx improves H1 by 0.989\% on average and reaches near-zero change at H32. The partition penalty appears at the initial prediction, whereas the small benefit of proximal regularization fades during rollout.

\FloatBarrier

\subsection{Algorithm behavior under fixed trajectory partitions (RQ3)}
\label{sec:diagnosis}
\textbf{RQ3: How do existing algorithms respond to trajectory heterogeneity under the fixed benchmark protocol?} We analyze three properties of the algorithm updates: their magnitude across rounds, the activation of extrapolation, and the sensitivity of cached updates to the trajectory partition.
\subsubsection{Update scales under the same temporal partition}
Update magnitudes differ substantially across methods using the same trajectory partition. We quantify this difference by dividing each server-update L2 norm by the paired FedAvg norm and taking the median across 32 disease--partition configurations at each round (Table~\ref{tab:updates}).

\begin{table}[!tbp]
\centering\small\papertablesetup
\caption{Update behavior behind the algorithm scores. Norm ratios are medians of within-configuration method/FedAvg ratios, calculated separately at each round. Own R5/R1 is the median ratio of each method's round-5 to round-1 norm, expressed as a percentage. Error changes average paired final scores.}
\label{tab:updates}
\begin{adjustbox}{max width=\textwidth}
\begin{tabular}{l||rrrr}
\papertoprule
\paperheadrow
\textbf{Method} & \textbf{Round 1 ratio} & \textbf{Round 5 ratio} & \textbf{Own R5/R1 (\%)} & \textbf{Error change (\%)}\\
\papergrouprule
\paperbaselinerow FedAvg & 1.0000 & 1.0000 & 5.74 & +0.00\\
\paperaltrow FedProx & 0.3871 & 0.8391 & 12.15 & -0.56\\
FedExP & 1.0000 & 1.1005 & 6.19 & -0.04\\
\paperaltrow FedLWS & 0.9704 & 0.9918 & 5.81 & -0.04\\
FedAvgM & 1.0000 & 23.0621 & 131.09 & +13.04\\
\paperaltrow FedAdam & 0.7634 & 18.4125 & 128.36 & +19.24\\
FedVARP & 1.0224 & 5.6960 & 30.60 & +21.96\\
\paperaltrow FedCDA & 0.1022 & 0.4938 & 26.27 & +17.77\\
SCAFFOLD & 0.0070 & 0.0035 & 2.81 & +22.21\\
\paperaltrow FedMuon & 0.0023 & 0.0025 & 6.24 & +26.78\\
\paperbottomrule
\end{tabular}
\end{adjustbox}
\end{table}
\begin{figure}[!tbp]
\centering
\includegraphics[width=\textwidth]{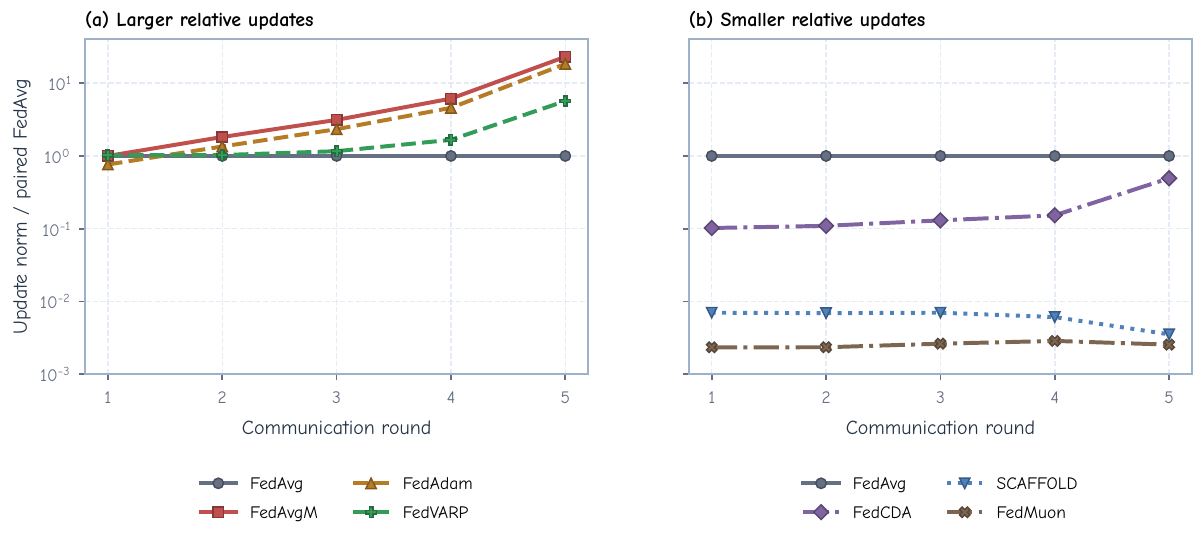}
\caption{Configured update rules produce different relative scales across rounds. Each point is the median of 32 within-configuration method/FedAvg update-norm ratios. Both panels use the same logarithmic scale and include FedAvg at one.}
\label{fig:updates}
\end{figure}
Figure~\ref{fig:updates} shows a wide spread in relative update scales. The within-method comparison in Table~\ref{tab:updates} explains how this spread develops: FedAvg's round-5 norm is only 5.74\% of its round-1 norm, whereas FedAvgM and FedAdam retain 131.09\% and 128.36\%, respectively. Their 23.06-fold and 18.41-fold round-5 ratios therefore reflect persistent server updates against a rapidly shrinking FedAvg update. FedVARP also decays more slowly, retaining 30.60\% of its initial norm. At the other extreme, SCAFFOLD and FedMuon already start at 0.0070 and 0.0023 times the FedAvg norm. High errors occur at both extremes: persistent large updates and consistently small updates. Update magnitude alone therefore does not explain prediction quality.

\paragraph{Server memory and changing trajectory samples.}
FedAvgM accumulates $v_r=0.9v_{r-1}+\delta_r$, where $\delta_r$ is the current aggregated client update. As the sampled patient shards change across rounds, momentum carries information from earlier trajectory segments into the current update. FedAvgM retains approximately its initial update magnitude at round five and has 13.04\% higher mean paired error than FedAvg. This combination motivates examining how well accumulated updates represent the currently sampled trajectory distribution.

\paragraph{Temporal relationships in local updates.}
Each update summarizes locally visible histories and future targets. RQ1 shows that these examples differ in both temporal extent and severity composition; RQ2 shows that improvements at the first predicted step can disappear during rollout. Together, these findings motivate learning representations that preserve temporal relationships across ownership boundaries and retain their predictive value at longer horizons.

\subsubsection{Mechanism activation across trajectory partitions}
\label{sec:fedexpactivation}
FedExP frequently follows the FedAvg update because extrapolation remains inactive. Across its 32 runs, the recorded extrapolation factor equals one in 119 of 160 rounds (74.4\%). Table~\ref{tab:fedexpactivation} shows that all 40 OASIS-30 rounds lie at this bound. Across all settings, ten runs have a factor of one in every round; their final reported nMAE values exactly equal the paired FedAvg values.

\begin{table}[!tbp]
\centering\small\papertablesetup
\caption{FedExP extrapolation by trajectory partition. Each partition contains eight disease runs with five rounds. A factor of one gives the FedAvg weighted update.}
\label{tab:fedexpactivation}
\begin{adjustbox}{max width=\textwidth}
\begin{tabular}{l||rrr}
\papertoprule
\paperheadrow
\textbf{Partition} & \textbf{Rounds at factor 1} & \textbf{Runs always at factor 1} & \textbf{Largest factor}\\
\papergrouprule
SOFA-30 & 30/40 & 1/8 & 1.262\\
\paperaltrow SOFA-50 & 24/40 & 0/8 & 1.372\\
OASIS-30 & 40/40 & 8/8 & 1.000\\
\paperaltrow OASIS-50 & 25/40 & 1/8 & 1.387\\
\paperbottomrule
\end{tabular}
\end{adjustbox}
\end{table}

In the remaining 22 FedExP runs, extrapolation activates in at least one round and reaches a maximum factor of 1.387, while final changes stay within $\pm1\%$ of FedAvg. The near-zero average gain thus combines inactive extrapolation with weak effects when extrapolation is active.

\begin{figure}[!tbp]
\centering
\includegraphics[width=\textwidth]{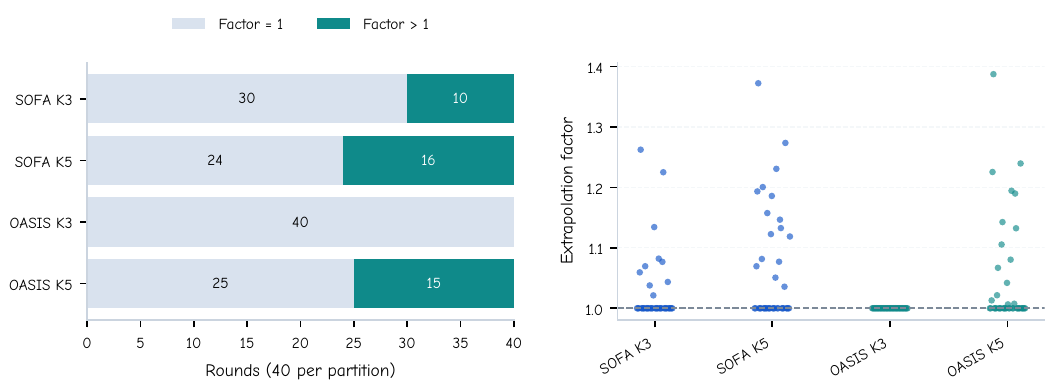}
\caption{FedExP activation under the benchmark protocol. Left: rounds at the factor-one floor versus active extrapolation. Right: all 40 round factors per partition; horizontal offsets separate observations for display.}
\label{fig:fedexp_activation}
\end{figure}

Figure~\ref{fig:fedexp_activation} shows the activation counts and all recorded factors by partition.

\subsubsection{Partition sensitivity of cached updates}
\begin{figure}[!tbp]
\centering
\includegraphics[width=\textwidth]{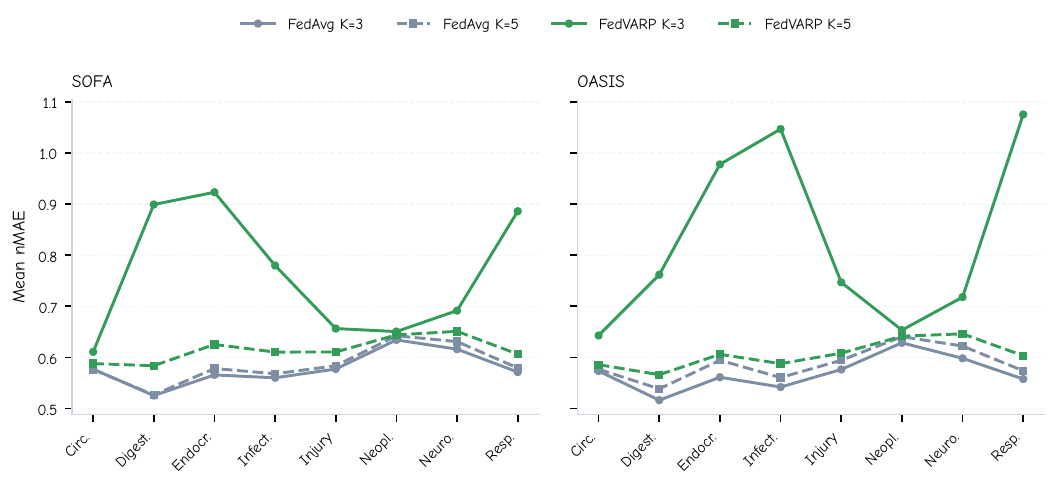}
\caption{Partition-dependent behavior of cached updates. FedVARP and FedAvg are compared within each disease under the same protocol. Lines connect six-horizon mean nMAE values across disease cohorts; solid and dashed lines denote three and five severity bands.}
\label{fig:cached_update_partition}
\end{figure}

Figure~\ref{fig:cached_update_partition} compares FedVARP and FedAvg across all eight diseases under the four benchmark partitions. FedVARP's mean paired degradation is 32.89\% under SOFA-30 and 5.15\% under SOFA-50; under OASIS, the corresponding values are 46.70\% and 3.12\%. The FedVARP penalty is much smaller with five severity bands, while FedAvg's own mean error rises under the finer partition. Finer ownership thus changes the value of the update cache as well as the difficulty of local prediction.

FedVARP reuses stored updates from previously participating clients. A round's server update can therefore combine information generated from different client mixtures and different global-model states. The large change in its penalty across partitions motivates tracking both the age of a cached update and the trajectory segments it represents.

\noindent\textbf{Finding 3: trajectory heterogeneity induces distinct algorithm responses under the same protocol.} Server methods exhibit different update persistence, FedExP frequently leaves extrapolation inactive, and FedVARP's degradation depends strongly on the partition. These behaviors connect the algorithm scores to how each update rule combines experience from heterogeneous trajectory segments.

\FloatBarrier

\subsection{Discussion and research opportunities}\label{sec:discussion}
\begin{figure}[!tbp]
\centering
\includegraphics[width=\textwidth]{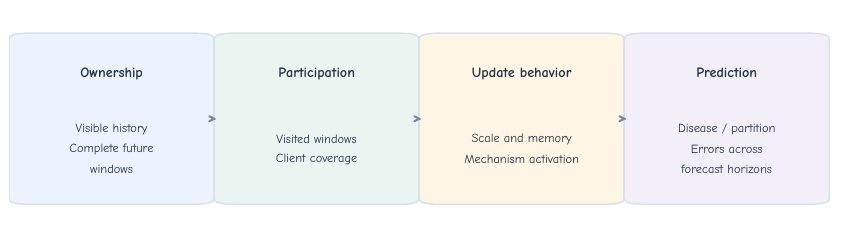}
\caption{Evidence chain for the fixed benchmark: temporal ownership determines visible context, participation determines its coverage, update behavior characterizes algorithm responses, and horizon-specific prediction measures outcomes.}
\label{fig:evidence_chain}
\end{figure}

Figure~\ref{fig:evidence_chain} summarizes the study's evidence chain under the fixed benchmark protocol. Ownership determines which history and future windows remain locally visible; participation determines which of those windows enter training; update behavior characterizes how the algorithms combine the resulting experience; and horizon-specific prediction measures the outcome.

\paragraph{Trajectory heterogeneity changes both support and composition.}
Only 7.55\%--21.36\% of pooled-available H32 windows have a locally complete anchor visited during training. Long-window availability also shifts the severity composition of the accessible examples. Finer partitions reduce retention in every cohort and accompany higher FedAvg error in 15 of 16 paired comparisons. These properties explain why transition counts alone provide an incomplete description of a federated world-model dataset.

\paragraph{Algorithm mechanisms respond differently to the same setting.}
FedProx's gains concentrate at short horizons, FedExP often reproduces weighted averaging, and FedVARP changes markedly across partitions. Update persistence and activation provide concrete descriptions of these responses. Future cross-time methods can build on this evidence by preserving temporal relationships across ownership boundaries and tracking which trajectory segments are represented in server-side memory~\cite{jhunjhunwala2022fedvarp}.

\paragraph{Data access and reproducibility.}
The benchmark uses credentialed-access MIMIC-IV data~\cite{johnson2024mimiciv31}. Construction specifications, field dictionaries, protocols, aggregate results, and analysis scripts support reconstruction and comparison.

%% file: section/conclusion.tex
\section{Conclusion}
Our study defines a cross-time benchmark for federated world-model learning and evaluates existing algorithms on clinical trajectories. The results reveal limited long-window coverage, small horizon-dependent gains, and partition-dependent algorithm behavior. These findings motivate federated learning methods that jointly account for temporal access, client participation, and the combination of trajectory-specific updates.